\documentclass{article}

\usepackage{arxiv}

\usepackage[utf8]{inputenc} % allow utf-8 input
\usepackage[T1]{fontenc}    % use 8-bit T1 fonts
\usepackage{hyperref}       % hyperlinks
\usepackage{url}            % simple URL typesetting
\usepackage{booktabs}       % professional-quality tables
\usepackage{amsfonts}       % blackboard math symbols
\usepackage{nicefrac}       % compact symbols for 1/2, etc.
\usepackage{microtype}      % microtypography
\usepackage{lipsum}
\usepackage{authblk}
\usepackage{graphicx}
\usepackage{subfigure}
\usepackage{mathtools}

\usepackage[skins,theorems]{tcolorbox}
\tcbset{highlight math style={enhanced,
  colframe=red,colback=white,arc=0pt,boxrule=1pt}}

\title{DisCoRL: Continual Reinforcement Learning via Policy Distillation}
\date{\vspace{-5ex}}

\author[*,1]{Ren\'e Traor\'e}
\author[*,1,3]{Hugo Caselles-Dupr\'e}
\author[*,1,2]{Timoth\'ee Lesort}
\author[1]{Te Sun}
\author[1]{Guanghang Cai}
\author[1]{Natalia D\'iaz-Rodr\'iguez}
\author[1]{David Filliat}% <-this % stops a space
\affil[1]{Flowers Team (ENSTA Paris, Institute Polytechnique de Paris \& INRIA).}%
\affil[2]{Thales, Theresis Laboratory.}%
\affil[3]{AI Lab, Softbank Robotics Europe}%
\affil[*]{Equal contribution.}% <-this % stops a space

\begin{document}
\maketitle

%===============================================================================

\begin{abstract}

In multi-task reinforcement learning there are two main challenges: at training time, the ability to learn different policies with a single model; at test time, inferring which of those policies applying without an external signal. In the case of continual reinforcement learning a third challenge arises: learning tasks sequentially without forgetting the previous ones. In this paper, we tackle these challenges by proposing DisCoRL, an approach combining state representation learning and policy distillation. We experiment on a sequence of three simulated 2D navigation tasks with a 3 wheel omni-directional robot. Moreover, we tested our approach's robustness by transferring the final policy into a real life setting. The policy can solve all tasks and automatically infer which one to run.

\end{abstract}

%%%%%%%%%%%%%%%%%%%%%%%%%%%%%%%%%%%%%%%%%%%%%%%%%%%%%%%%%%%%%%%%%%%%%%%%%%%%%%%%
%%%%%%%%%%%%%%%%%%%%%%%%%%%%%%%%%%%%%%%%%%%%%%%%%%%%%%%%%%%%%%%%%%%%%%%%%%%%%%%%
%%
%                                    Introduction
%%
%%%%%%%%%%%%%%%%%%%%%%%%%%%%%%%%%%%%%%%%%%%%%%%%%%%%%%%%%%%%%%%%%%%%%%%%%%%%%%%%
%%%%%%%%%%%%%%%%%%%%%%%%%%%%%%%%%%%%%%%%%%%%%%%%%%%%%%%%%%%%%%%%%%%%%%%%%%%%%%%%

\section{Introduction}

% autonomous agent vs machine learning
An autonomous agent should be able to learn and exploit its knowledge in any situation all along its life. In a sequence of learning experiences, it should therefore be able to build representations and skills that can be reactivated and reused later. Machine learning is a research area that addresses the problem of learning automatically from any agent experiences. Nevertheless several challenges still block the way toward a fully autonomous agent. In particular, we focus on the situations when the agent needs to learn skills sequentially into a single model and use them independently afterwards.

% multi task and continual 
This challenge is partially addressed by a sub-domain of machine learning called multi-task learning. Multi-task learning \cite{Caruana97} studies how to optimize several problems {\it simultaneously} with a single model. However, when those problem can not be optimized at the same time and have to be learned sequentially, we identify the learning setting as a continual learning problem \cite{LesortLomonacoDiaz19}.
In this paper, we propose to address a continual learning problem of reinforcement learning (RL). In this continual learning setting, each learning experience is called a task and a task solution is a policy.

\begin{figure}[ht]
\centering
    \includegraphics[width=0.6\textwidth]{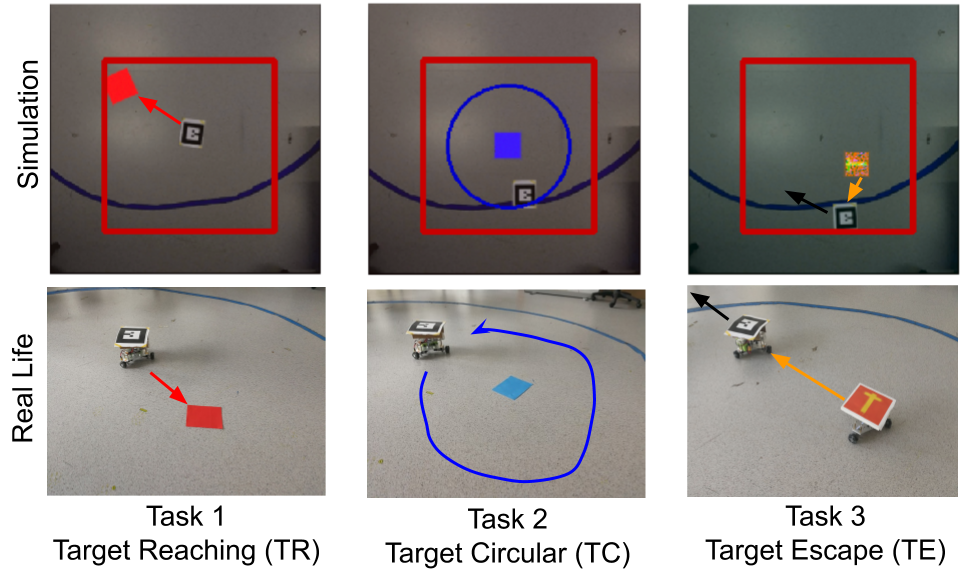} 
    \caption{Image of the three tasks, in simulation (top) and in real life (bottom) sequentially experienced. Learning is performed in simulation, real life is only used at test time. }
    \label{fig:real-life-tasks}
\end{figure}

% presentation setting
The goal is to propose a learning setting compatible with a real autonomous agent. To this end, we propose three simulated robotics tasks and propose an approach that will solve such tasks sequentially. At each task, the agent should learn a policy based on a reward function and a RL algorithm.

% RL in this context
RL is a popular framework to learn robot controllers that also has to face the CL challenges.
In order to fit reinforcement learning into a continual setting, we use a method called policy distillation \cite{rusu2015policy} that allows to transfer several policies learned sequentially into a single model.
% Real life
To validate our approach, we evaluate the final results on the three simulated learning setting but also in a real life setting similar to the simulation (Figure \ref{fig:real-life-tasks}). 
% no task index
It is important to note that, at test time, the agent does not have access to a task label to determine which policy to run, and thus, it needs to figure it out by itself from its observations.

% Contribution
Our contribution is to propose \textbf{DisCoRL} (\textit{Distillation for Continual Reinforcement learning}): a modular, effective and scalable pipeline for continual RL. This pipeline uses policy distillation for learning without forgetting, without access to previous environments, and without task labels. Our results show that the method is efficient and learns policies transferable into real life scenarios.

%Plan. % if space:
The article is structured as follows. Sec.~\ref{ref:relatedwork} introduces related work,  Sec. \ref{ref:methods} details the methods utilized, Sec. \ref{ref:experiments} describes the robotics setting and  tasks, Sec. \ref{ref:Results} presents the experiments performed, and Sec. \ref{ref:discussion} concludes with future insights from our experiments.

%%%%%%%%%%%%%%%%%%%%%%%%%%%%%%%%%%%%%%%%%%%%%%%%%%%%%%%%%%%%%%%%%%%%%%%%%%%%%%%%
%%%%%%%%%%%%%%%%%%%%%%%%%%%%%%%%%%%%%%%%%%%%%%%%%%%%%%%%%%%%%%%%%%%%%%%%%%%%%%%%
%%
%                                    Related work
%%
%%%%%%%%%%%%%%%%%%%%%%%%%%%%%%%%%%%%%%%%%%%%%%%%%%%%%%%%%%%%%%%%%%%%%%%%%%%%%%%%
%%%%%%%%%%%%%%%%%%%%%%%%%%%%%%%%%%%%%%%%%%%%%%%%%%%%%%%%%%%%%%%%%%%%%%%%%%%%%%%%

\section{Related work}
\label{ref:relatedwork}

\quad \textbf{Multi-task RL: }
The objective of Multi-task learning (MTL) \cite{Caruana97} is to learn several tasks simultaneously; generally by training tasks in parallel with an unique model.
Therefore, multi-task RL aims at constructing one single policy that can solve a number of different tasks. Note how in classification this problem is quite simple, as data from all tasks just have to be shuffled randomly and can then be learned all together at once. However, in RL environments, data is sampled on sequences that can not be shuffled randomly with all other environments because the environments are not accessible simultaneously. Learning multiple tasks at once is thus more complicated.

%policy distillation
Policy distillation \cite{rusu2015policy} can be used to merge different policies into one single module/network.  This approach uses two models, a trained policy (the teachers) to annotate data with soft-annotations, and a model to learn from the former (the student). The student is trained in a supervised manner with the soft-labels. The soft-annotation is supposed to help the student to learn faster than the teacher did \cite{Furlanello18}.
Policy distillation can be used then to learn several policies separately and simultaneously, and distill them into a single model as in the distral algorithm \cite{teh2017distral}. In our approach, we also use distillation but we do not keep the teacher model, we just label a set of data and then delete the teacher. Furthermore, tasks are learned sequentially, and not simultaneously.
Other approaches such as SAC-X \cite{riedmiller2018learning} or HER \cite{andrychowicz2017hindsight} take advantage of Multi-task RL by learning auxiliary tasks in order to help learning a main task. 
%curious
This approach is extended in the CURIOUS algorithm \cite{colas2018curious}. It selects tasks to be learned that improve an absolute learning progress metric the most.

\quad \textbf{Continual Learning: }
Continual learning (CL) is the ability of a model to learn new skills without forgetting previous knowledge \cite{LesortLomonacoDiaz19}. Continual learning is in many ways similar to multi-task learning but past task can not be replayed.
In our context, it means learning several tasks sequentially and being able to solve any of the learned tasks at the end of the sequence.  

Most CL approaches can be classified into four main methods that differ in the way they handle the memory from past tasks. The first method, referred to as \textit{rehearsal}, keeps samples from previous tasks 
\cite{rebuffi2017icarl, nguyen2017variational}.
The second approach consists of applying \textit{regularization}, either by constraining weight updates in order to maintain knowledge from previous tasks 
\cite{kirkpatrick2017overcoming, Zenke17, Maltoni18}, or by keeping an old model in memory and distilling knowledge \cite{hinton2015distilling} into it later to remember \cite{li2018learning, schwarz2018progress}.
The third category of strategies, \textit{dynamic network architectures}, maintains past knowledge thanks to architectural modifications while learning \cite{rusu2016progressive, fernando2017pathnet, li2018learning, fernando2017pathnet}.
The fourth method is \textit{generative replay} \cite{shin2017continual, lesort2018generative, wu2018memory, lesort2018marginal}, where a generative model is used as a memory to produce samples from previous tasks. 

In the context of continual reinforcement learning, several approaches have been proposed, such as the use of \textit{Progressive Nets} in \cite{Rusu16Sim}, \textit{EWC} \cite{kirkpatrick2017overcoming}, \textit{Progress And Compress} (P\&C) \cite{schwarz2018progress}, or \textit{CRL-Unsup} \cite{Lomonaco19}. 
However they either need a task indicator at test time to choose which policies  to run or, they have some hyper-parameter difficult to tune during a continual learning training, such as the importance of the Fisher information matrix in EWC. Our method does not add any new hyper-parameter to tune during the sequence of tasks and does not need a task label at test time. 

\quad \textbf{Reinforcement learning in Robotics: }
Applying RL to real-life scenarios such as robotics is a major challenge that has been studied widely. 

One of the major problems in this setting is that sampling data and a fortiori learning is costly. Therefore sample efficiency and stability in learning are highly valuable.
One common approach to reduce training cost, is training policies in simulation and then deploying them in real-life hoping that they will successfully transfer, considering the gap in complexity between simulation and the real world. Such approaches are termed \textit{Sim2Real} \cite{Golemo19}, and have been successfully applied \cite{christiano2016transfer, matas2018sim} in many scenarios. One of these approaches is Domain Randomization \cite{tobin2017domain}, which we use in this paper. This technique trains policies in numerous simulations that are randomly different from each other (different background, colors, etc.). Using this technique, the transfer to real life is easier.

Another method we also exploit is to first learn a state representation \cite{Lesort18} to compress the observation into a low dimensional embedding and secondly learn the policy on top of this representation. This method helps to improve sample efficiency and stability of RL algorithms \cite{raffin2019decoupling} and thus can make them directly applicable in real life.

Others have tried to train a policy directly on real robots, facing the hurdle of the lack of sample efficiency of RL algorithms. SAC-X \cite{riedmiller2018learning} is one example that takes advantage of multi-task learning to improve efficiency, by simultaneously learning the policy and a set of auxiliary tasks to explore its observation space - in search for sparse rewards of the externally defined target task.

In the literature, most approaches focus on the single-task or simultaneous multi-task scenario. In this paper, we attempt to train a policy on several tasks sequentially and deploy it in real life by combining policy distillation, training in simulation and state representation learning.

%%%%%%%%%%%%%%%%%%%%%%%%%%%%%%%%%%%%%%%%%%%%%%%%%%%%%%%%%%%%%%%%%%%%%%%%%%%%%%%%
%%%%%%%%%%%%%%%%%%%%%%%%%%%%%%%%%%%%%%%%%%%%%%%%%%%%%%%%%%%%%%%%%%%%%%%%%%%%%%%%
%%
%                                    Methods
%%
%%%%%%%%%%%%%%%%%%%%%%%%%%%%%%%%%%%%%%%%%%%%%%%%%%%%%%%%%%%%%%%%%%%%%%%%%%%%%%%%
%%%%%%%%%%%%%%%%%%%%%%%%%%%%%%%%%%%%%%%%%%%%%%%%%%%%%%%%%%%%%%%%%%%%%%%%%%%%%%%%

\section{Methods}
\label{ref:methods}

In this section we present our approach towards continual reinforcement learning for a sequence of vision based tasks. We assume that observations visually allow to recognize the current task from other tasks. We first explain how we learn a single task by combining SRL and RL, then how each task is incorporated in the continual learning pipeline. Finally, we present how we evaluate the full pipeline. 

\subsection{Learning one task}
\label{subsec:oneTask}

Each task $i$ is solved by first learning a state representation encoder $E_i$ in order to compress input images into a representation of the important underlying factor of variation.
% why using srl
This step allows to reduce the input space for the reinforcement learning algorithm and makes it learn more efficiently \cite{raffin2019decoupling}.
To train this encoder, as shown in Fig. \ref{fig:overview} (left), we sample data from the environment $Env_i$ 
with a random policy. We call this dataset $D_{R,i}$.
% learning SRL_t
$D_{R,i}$ is then used to train the SRL model composed of an \textit{inverse model} and an \textit{auto-encoder}. The inverse model is trained to predict the action $a_t$ that led to transition from state $s_t$ to $s_{t+1}$, both extracted from respective observations $o_t$ and $o_{t+1}$ by the auto-encoder using $E_i$. The auto-encoder is additionally trained to reconstruct the observations from the encoded states. The architecture is motivated by the results from \cite{raffin2019decoupling}, and illustrated in the Appendix F.

% learning pi_t
Once the SRL model is trained, we use its encoder $E_i$ to provide features as input of a policy $\pi_i$ trained using RL. We also experimented to learn the policy directly in the raw pixel space but, as shown in \cite{raffin2019decoupling}, it was less sample efficient. 

Once $\pi_i$ is learned, we use it to generate sequences of on-policy observations with associated actions, which will eventually be used for distillation (Fig. \ref{fig:overview}, right). We call this the distillation dataset $D_{\pi_i}$. We generate $D_{\pi_i}$ the following way: we randomly sample a starting position and then let the agent generate a trajectory. At each step we save both the observation and associated action probabilities. We collect the shortest sequences maximizing the reward for an episode (see Section \ref{subsec:evaluation}).% N early stopping? any 
We also experiment to generate $D_{\pi_i}$ with a regular sampling and a random policy but annotated with $\pi_i$ to compare results, as detailed in section \ref{sec:sampling-strategies-section}.

From each task we only keep dataset $D_{\pi_i}$. As soon as we change task, $D_{R,i}$ and $Env_i$ are not available anymore. $D_{\pi_i}$ is split into a training set and a validation set.

\subsection{Learning continually}
\label{subsec:continual_learning}

\begin{figure*}
\centering
\includegraphics[width=0.7\textwidth]{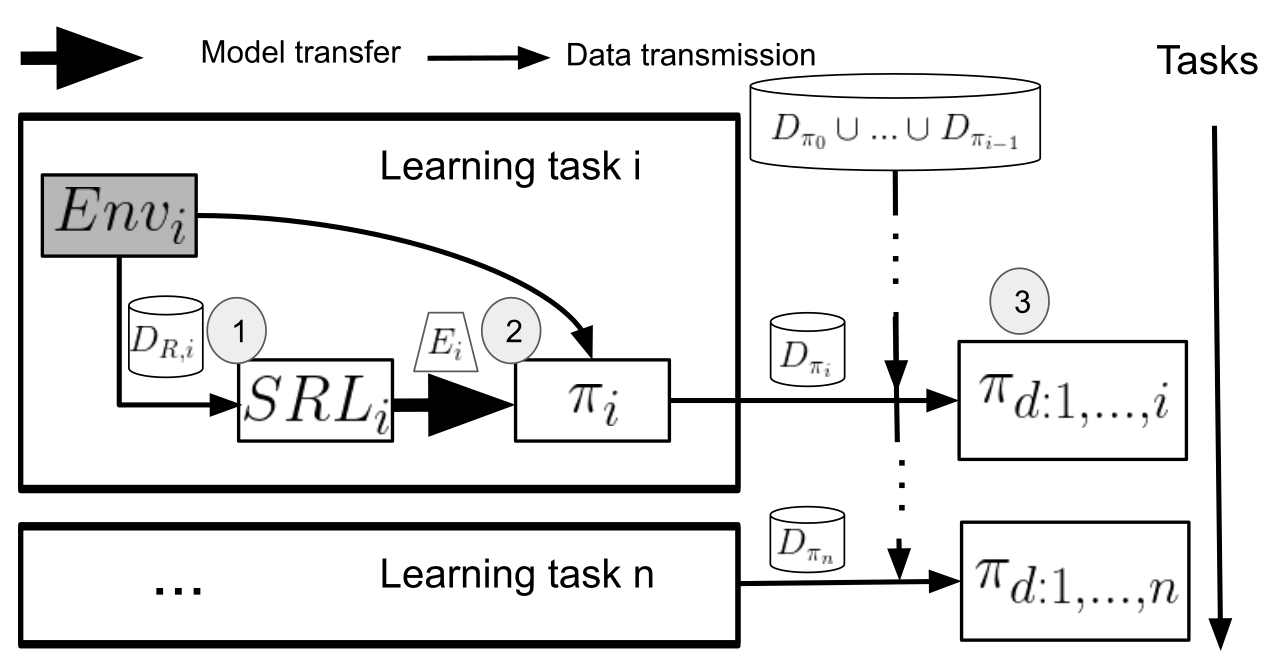}
\caption{Overview of our full pipeline for Continual Reinforcement Learning. White cylinders are for datasets, gray squares for environments, and white squares for learning algorithms, whose name corresponds to the model trained. 
Each task $i$ is learned sequentially and independently by first generating a dataset $D_{R,i}$ with a random policy to train a state representation with an encoder $E_i$ with an SRL method (1), then we use $E_i$ and the environment to learn a policy $\pi_i$ in the state space (2). Once trained, $\pi_i$ is used to create a distillation dataset $D_{\pi_i}$ that acts as a memory of the learned behaviour. 
All policies are finally compressed into a single policy $\pi_{d:{1,..,i}}$ by merging the current dataset $D_{\pi_i}$ with datasets from previous tasks $D_{\pi_1} \cup ... \cup D_{\pi_{i-1}}$ and using distillation (3).} 

\label{fig:overview}
\end{figure*}

To learn continually we adapt policy distillation \cite{rusu2015policy} to a continual learning setting. The distillation consists in training a student policy to imitate a teacher policy. In our case, a student model
learn from a teacher policy the action probability associated to each observation. 
Each dataset $D_{\pi_i}$ allows to distill the policy $\pi_i$  (the \textbf{teacher} model) into a new network $\pi_{d:i}$ (the \textbf{student} model). 
In classic distillation, both data and models need to be saved, however saving just soft-annotated data is a lighter solution adapted to a continual setting.

 With the aggregation of several distillation datasets $D_{\pi_i}$, we can distill several policies into the same network that can achieve all tasks (Fig.\ref{fig:overview}, bottom right). By extension of the previous nomenclature, we denote $\pi_{d:1,..,n}$ a model where  policies $\pi_1$ ... $\pi_n$ have been distilled in. 
 When distilling all policies into the student, we select our best models with early stopping, and test later in simulation and in real life settings.

 Since we assume that observations visually allow to recognize the current task, $\pi_{d:1,..,n}$ is able to choose the right action for the current task without a task indicator.

The method, termed \textit{DisCoRL} for \textit{Distillation for Continual Reinforcement learning}, allows to learn continually several policies while minimizing forgetting. Regarding scalability, saving data from all past experiments may not look ideal if there is a high number of tasks. However, this solution is highly effective for remembering
and letting the reinforcement learning algorithm be absolutely free to learn a new policy without regularization. It is worth mentioning that RL is the real bottleneck in the whole process: Dataset $D_{\pi_i}$ contains approximately 10k samples per task, which allows to perform the distillation quickly, relative to how long and computationally expensive RL is (few minutes needed to learn $\pi_{d:i}$ while several hours are needed to learn $\pi_i$). Thus, in this context, it is better not to curb RL with regularization. Indeed, as explained in Section \ref{subsec:neg_res}, we tried several regularization based approaches that were not successful.

Within the Continual Learning framework for robotics \cite{LesortLomonacoDiaz19}, our setting falls within the category of Multi Task learning scenario (MT) with a NIC (New Instances and New Concepts) content update type for each task. Our approach can be classified into the rehearsal family of approaches where memory is saved as data points.

\subsection{Evaluation}
\label{subsec:evaluation}

% evaluation in sim
The first evaluation is the performance of the final policy on the simulated environment.
This evaluation can then be compared with the performance of each teacher policy.
% evaluation in real
For the second evaluation we test if the policy is robust to the reality gap and can be adapted into a real life scenario. The simulation is voluntary close the real life setting but the reality gap is notoriously problematic. 

To get an insight on the evolution of the distilled model, we also save distillation datasets at different checkpoints while learning each tasks. By distilling and evaluating at several time steps, we assess catastrophic forgetting on previous task when finetuning on new a task (See Appendix D).

%%%%%%%%%%%%%%%%%%%%%%%%%%%%%%%%%%%%%%%%%%%%%%%%%%%%%%%%%%%%%%%%%%%%%%%%%%%%%%%%
%%%%%%%%%%%%%%%%%%%%%%%%%%%%%%%%%%%%%%%%%%%%%%%%%%%%%%%%%%%%%%%%%%%%%%%%%%%%%%%%
%%
%                                    Experimental setup
%%
%%%%%%%%%%%%%%%%%%%%%%%%%%%%%%%%%%%%%%%%%%%%%%%%%%%%%%%%%%%%%%%%%%%%%%%%%%%%%%%%
%%%%%%%%%%%%%%%%%%%%%%%%%%%%%%%%%%%%%%%%%%%%%%%%%%%%%%%%%%%%%%%%%%%%%%%%%%%%%%%%

\section{Experimental setup}
\label{ref:experiments}

We apply our approach to learn continually three 2D navigation tasks applicable in real life.
The software related to our experimental setting is available online
\footnote{\url{https://github.com/kalifou/robotics-rl-srl}}. \\

\subsection{Robotic setup}

The experiments consists of 2D navigation tasks using a 3 wheel omni-directional robot similar to the 2D mobile navigation in \cite{raffin2018s}. The input image is a top-down view of the floor and the robot is identified by a black QR code. The room where the real-life robotic experiments are performed is lighted by surroundings windows and artificial illumination and is subject to illumination changes depending on the weather and time of the day. 
The robot uses 4 high level discrete actions (move left/right, move up/down in a cartesian plane relative to the robot) rather than motor commands.

We simulate the experiment to increase sampling and learning speed. The simulation is performed by artificially moving the robot picture inside the background image according to the chosen actions.
We use domain randomization \cite{tobin2017domain} to improve the stability and facilitate transfer to the real world : during RL training, at each timestep, the color of the background is randomly changed.

\subsection{Continual learning setup}
\label{sec:tasks_presentation}

Our continual learning scenario is composed of three similar environments, where the robot is rewarded according to the associated task (Fig. \ref{fig:real-life-tasks}). 
In all environments, the robot is free to navigate for up to 250 steps, performing only discrete actions  
within the boundaries identified by a red line. Each task is associated to a visual target, which color depends on the task. This way, the controller can automatically infer which policy it needs to run and thus, does not need task labels at test time.

\quad \textbf{Task 1.} The task of environment 1 is named Target Reaching (TR). The robot gets at each timestep $t$ a positive reward $+1$ for reaching the target (red square), 
a negative reward $-1$ for bumping into the boundaries, and no reward otherwise. 

\quad \textbf{Task 2.} The task of environment 2 is named Target Circling (TC). The robot gets at each timestep $t$ a reward $R_t$ defined in Eq. \ref{eq:reward-circular-task} (where $z_t$ is the 2D coordinate position with respect to the center of the circle) designed for agents to learn the task of circling around a central blue tag. This reward is highest when the agent is both on the circle (red (first) square in Eq. \ref{eq:reward-circular-task}), and has been moving for the previous $k$ steps (blue, second square).
An additional penalty term of $-1$ is added to the reward function in case of bump with the boundaries (last, green square. A coefficient $\lambda=10$ is introduced to balance the behaviour.

\tcbset{reward-moving/.style={no shadow,colframe=blue, boxrule=0.5pt,frame style={opacity=0.25}}} 
\tcbset{reward-circular/.style={no shadow,colframe=red, boxrule=0.5pt,frame style={opacity=0.25}}} 
\tcbset{bump/.style={no shadow,colframe=green, boxrule=0.5pt,frame style={opacity=0.25}}} 
 
\begin{equation}
R_t =
\lambda *
\tcbhighmath[reward-circular]{(1 - \lambda (\|z_t\| -  r_{circle}) ^2)}
*  
\tcbhighmath[reward-moving]{\|z_t -z_{t-k}  \|_{2}^2}
+ 
\lambda ^ 2 *
\tcbhighmath[bump]{R_{t, bump}}
\label{eq:reward-circular-task}
\end{equation}

\quad \textbf{Task 3.} The task of environment 3 is named Target Escaping (TE). Robot A is being chased down by another robot B with an orange tag. Robot B is hard-coded to follow robot A, and robot A has to learn to escape using RL. Robot A gets at each timestep $t$ a reward of $+1$ if it's far enough from robot B, otherwise, if it is in the range of robot B, it gets a reward of $-1$. Additionally, robot A gets a negative reward $-1$ for bumping into the boundaries.

All RL tasks are learned with PPO2 \cite{Schulman17} and the same SRL model, as described in section \ref{subsec:oneTask}. We select the model architecture as in \cite{raffin2018s} for RL and SRL. All datasets size and characteristics are described in the Appendix. The input observations of all models are RGB images of size $224 * 224 * 3$.

%%%%%%%%%%%%%%%%%%%%%%%%%%%%%%%%%%%%%%%%%%%%%%%%%%%%%%%%%%%%%%%%%%%%%%%%%%%%%%%%
%%%%%%%%%%%%%%%%%%%%%%%%%%%%%%%%%%%%%%%%%%%%%%%%%%%%%%%%%%%%%%%%%%%%%%%%%%%%%%%%
%%
%                                    Results
%%
%%%%%%%%%%%%%%%%%%%%%%%%%%%%%%%%%%%%%%%%%%%%%%%%%%%%%%%%%%%%%%%%%%%%%%%%%%%%%%%%
%%%%%%%%%%%%%%%%%%%%%%%%%%%%%%%%%%%%%%%%%%%%%%%%%%%%%%%%%%%%%%%%%%%%%%%%%%%%%%%%

\section{Results}
\label{ref:Results}

We first present our design choices for the distillation process: loss functions and data sampling strategies. We then use these choices to present our main result: the distillation of three tasks continually into a single policy that can achieve the three tasks both in simulation and real-life. We provide a supplementary video of this policy deployed in real-life on the robot showing the successful behaviors. We also present the different strategies we tried but that did not work in our setting.

\subsection{Evaluation of distillation}
\label{sec:sampling-strategies-section}

\quad \textbf{Distillation strategies:} Distillation is done with a loss function that minimizes the difference between the student model's output and the teacher model's output for the same input. As in the policy distilation paper \cite{rusu2015policy}, we investigate variations of the loss function : Mean Squared Error loss ($\mathcal{L}_{MSE}(x,y) = \mathbb{E}[||x - y||_{2}^{2}]$), Kullback-Lieber divergence, and Kullback-Lieber divergence with temperature smoothing ($\mathcal{L}_{KL,\tau}(p|q) = \mathbb{E}[\text{softmax}(\frac{p}{\tau}) ln(\frac{\text{softmax}(\frac{p}{\tau})}{\text{softmax}(q)}))]$).

We run a performance comparison of the different losses by computing the mean normalized performance of a student policy trained to perform all three tasks (Tab.\ref{tab:distillation-losses}). Using the Kullback-Lieber divergence loss function with temperature smoothing with $\tau = 0.01$ is best, and optimizing the temperature parameter yields a small performance boost. This result is coherent with \cite{rusu2015policy} where they reach the same conclusion.

\begin{table}[h]
\centering
\begin{tabular}{lc}
\textbf{Distillation loss} & \textbf{Student performance ($\pm$ std)} \\ \hline
MSE & 0.71 ($\pm$ 0.22) \\ \hline
KL ($\tau = 1$) & 0.76 ($\pm$ 0.14) \\ \hline
KL ($\tau = 0.1$) & 0.68 ($\pm$ 0.18) \\ \hline
KL ($\tau = 0.01$) & \textbf{0.77 ($\pm$ 0.13)}
\end{tabular}
\caption[test]{Mean normalized performance\footnote{Normalization is done by reducing the episode reward between 0 and the max. possible performance to a reward in [0,1].} of a student policy trained with distillation using 4 different loss functions. The student policy is trained to perform all three tasks. Kullback-Lieber divergence with $\tau=0.01$ performs best.}
\label{tab:distillation-losses}
\end{table}

\quad \textbf{Data sampling strategies:}
We evaluate the effect of two different sampling strategies to create $D_{\pi_i}$ for policy distillation. Data sampling is a key component as the sampled dataset should be as small as possible but contain sufficient information for student model training. 
The strategies involved for data generation are: \\

\begin{figure}
\centering
\includegraphics[width=0.18\textwidth]{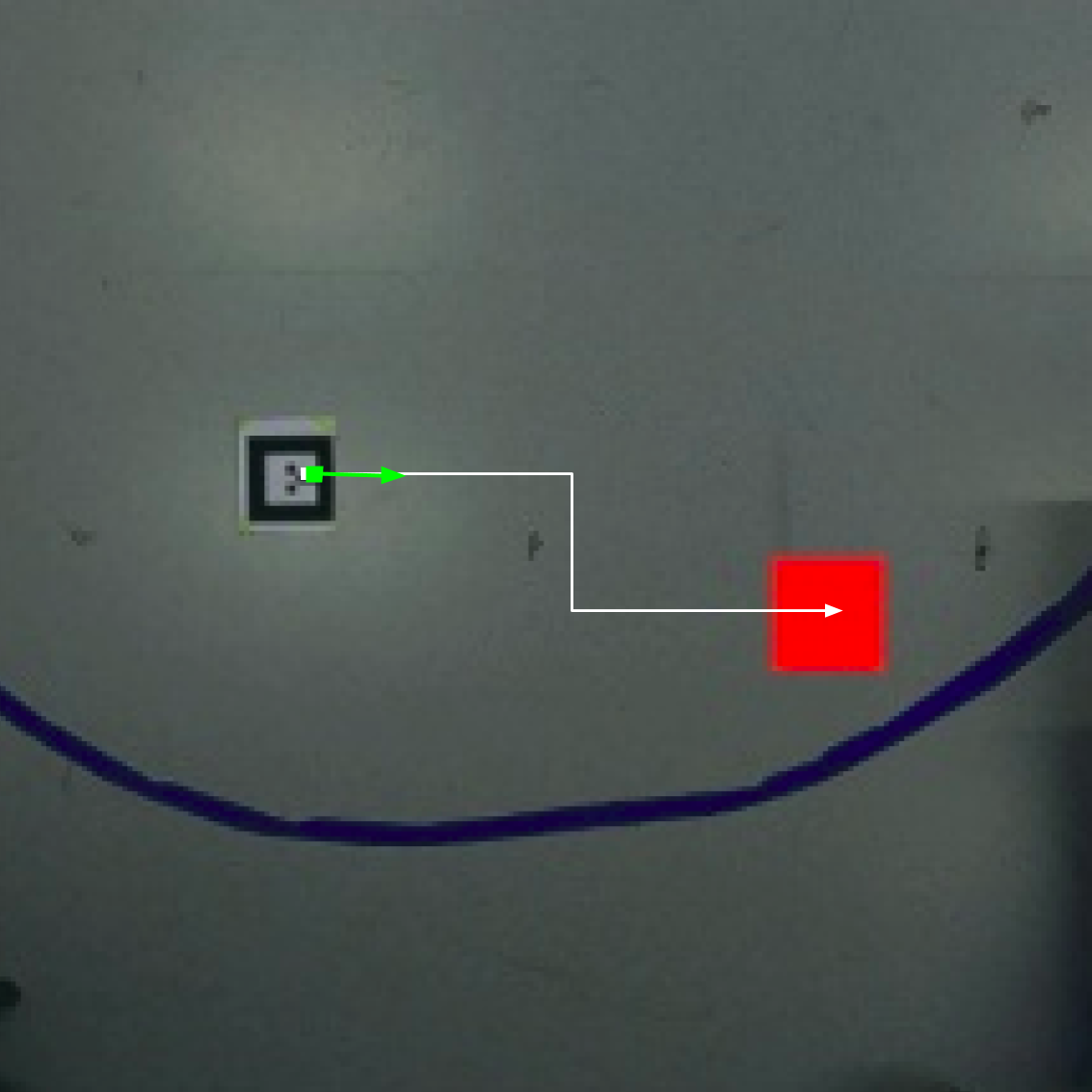}
\includegraphics[width=0.18\textwidth]{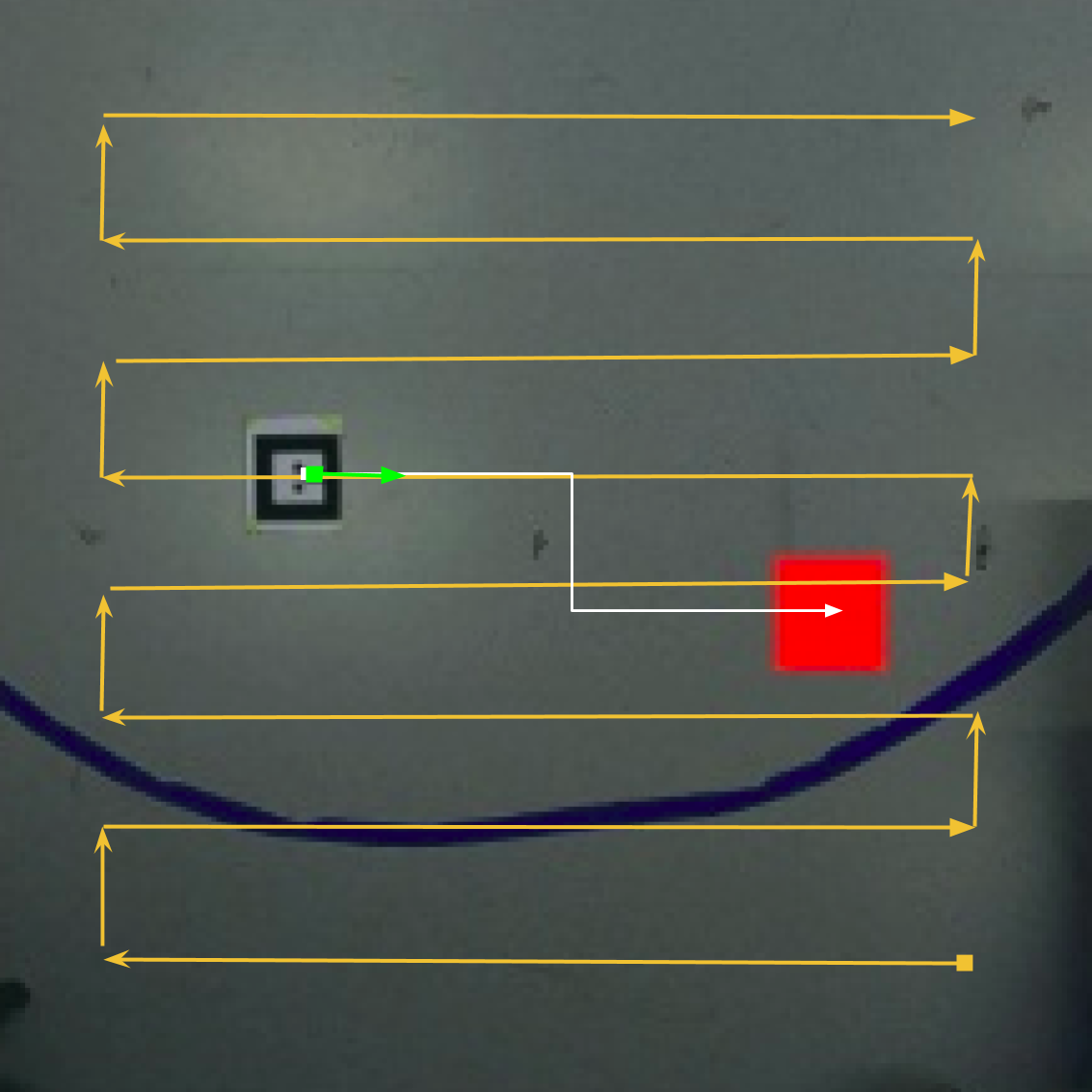}
\caption{}
\label{fig:data-generation-strategies}
\end{figure}
- \textit{On-policy generation (Fig.\ref{fig:data-generation-strategies}, top)}: We start an episode from a random point, then at each timestep $t$, we collect an observation $o_t$ and perform the action $a_{\pi_i, t}$ of the teacher policy. 
$D_{\pi_i}$ is thus composed of tuples ($o_t, p(a_{\pi_i, t} \mid o_t)$), with $p(a_{\pi_i, t} \mid o_t)$ the action probability associated to the action $a_{\pi_i, t}$ taken by the teacher, i.e., a \textit{soft label}, since we use the Kullback-Lieber divergence loss.

- \textit{Off-policy generation from a grid walker (Fig.\ref{fig:data-generation-strategies}, bottom)}: at each timestep $t$, we collect an observation $o_t$ by performing an action $a_{grid, t}$ of a \textit{grid walker} exhaustively exploring the space of the arena. However, for each $o_t$ we save the probability of action $p(a_{\pi_i, t} \mid o_t)$ that would have been taken by a teacher policy. $D_\pi$ is thus composed of tuples ($o_t, p(a_{\pi_i, t} \mid o_t)$).
The goal of this strategy is to provide a more exhaustive sampling of the space of robot positions.

\begin{figure}[h]
    \centering
    \includegraphics[width=0.34\linewidth]{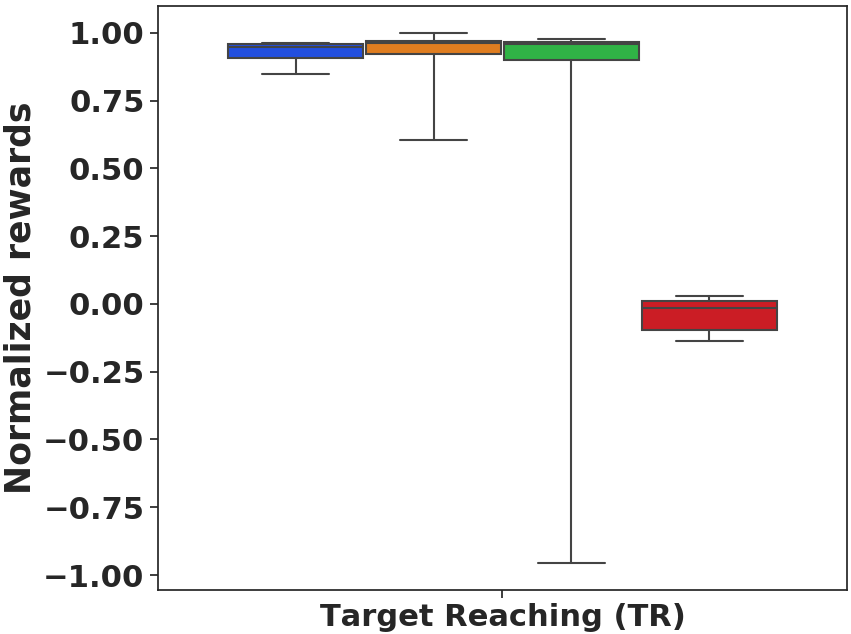}
    \includegraphics[width=0.32\linewidth]{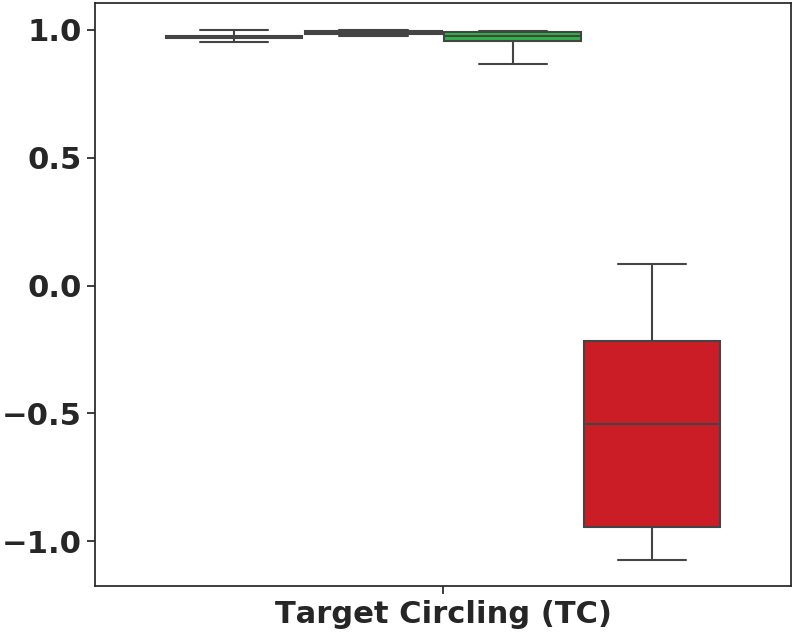}
    \includegraphics[width=0.32\linewidth]{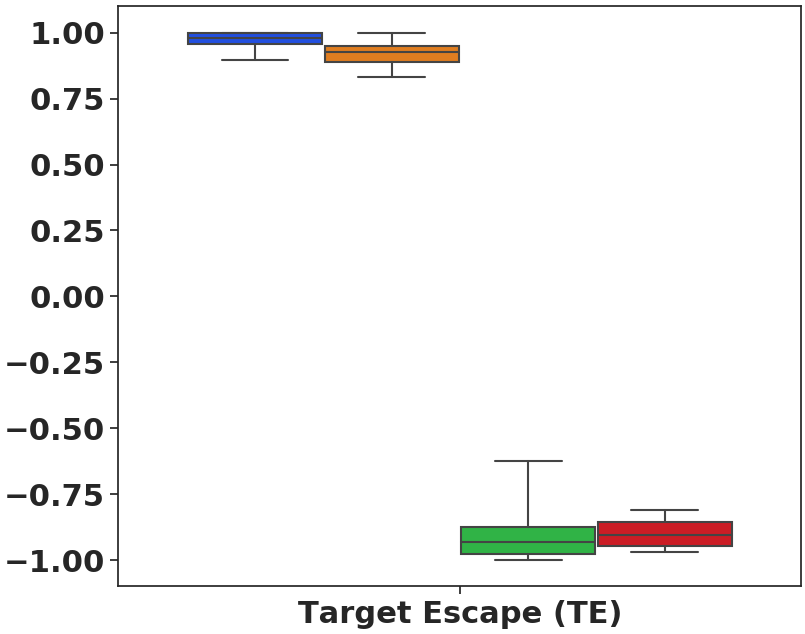}
    \includegraphics[width=.99\linewidth]{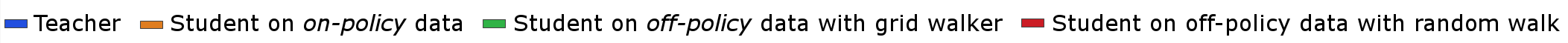}
    \caption{Efficiency (normalized rewards w.r.t the best teacher performance) of policies distilled on 8 seeds using various data generation strategies for each task separately. Each evaluated policy is distilled on 15k tuples of sampled observations and action probabilities, for 4 epochs (see criteria of stopping in section \protect\ref{subsec:continual_learning} and Appendix B).} 
    \label{fig:comparing-on-policy-strategies}
\end{figure}

Performance of policies distilled using such strategies (see Fig. \ref{fig:comparing-on-policy-strategies}) show that \textit{on-policy generation} (i.e., demonstrations) suffice to reproduce performance closed to those of teacher policies on every task individually, with reasonable stability. In particular cases, see Fig. \ref{fig:comparing-on-policy-strategies} for task TC, 
this strategy even provides a small boost in performances in the student policy over the teacher policy.

However, using \textit{off-policy data generation from a grid walker} for distillation results in either unstable or poorly performing policies, especially in tasks defined by a reward function requiring the agent to move actively (\textit{TC} task, blue part of eq. \ref{eq:reward-circular-task}) or anticipate the behaviour of another agent (TE task). 
In this case, the resulting policy reaches the performances of a lower-bound baseline obtained by distilling from trajectories of an untrained policy (see \textit{Student on off-policy data with a random walker} in fig. \ref{fig:comparing-on-policy-strategies}), i.e. from a policy with random weights with input in the raw pixels' space.

\subsection{Main result}

\begin{figure}[h]
    \centering
    \includegraphics[width=0.32\linewidth]{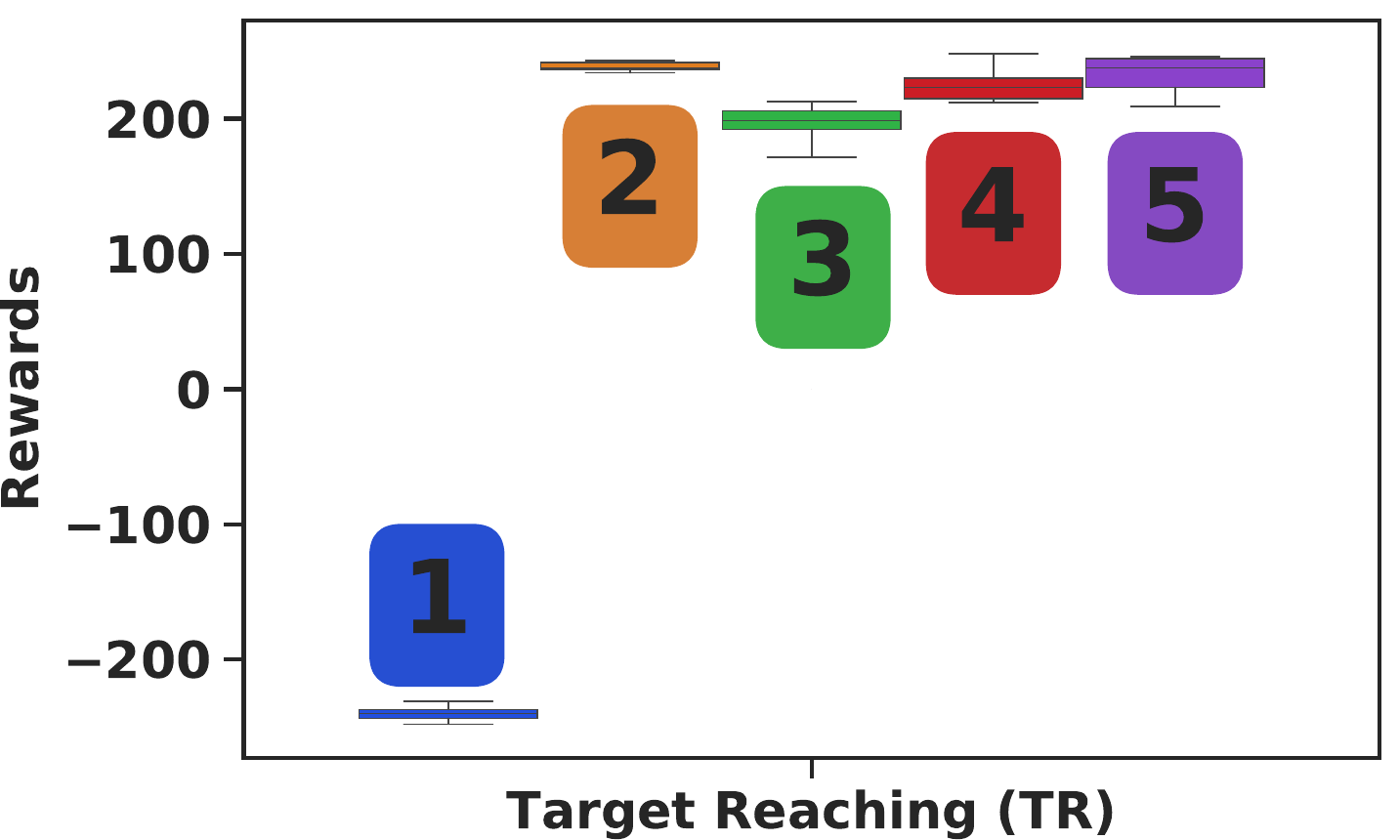}
    \includegraphics[width=0.32\linewidth]{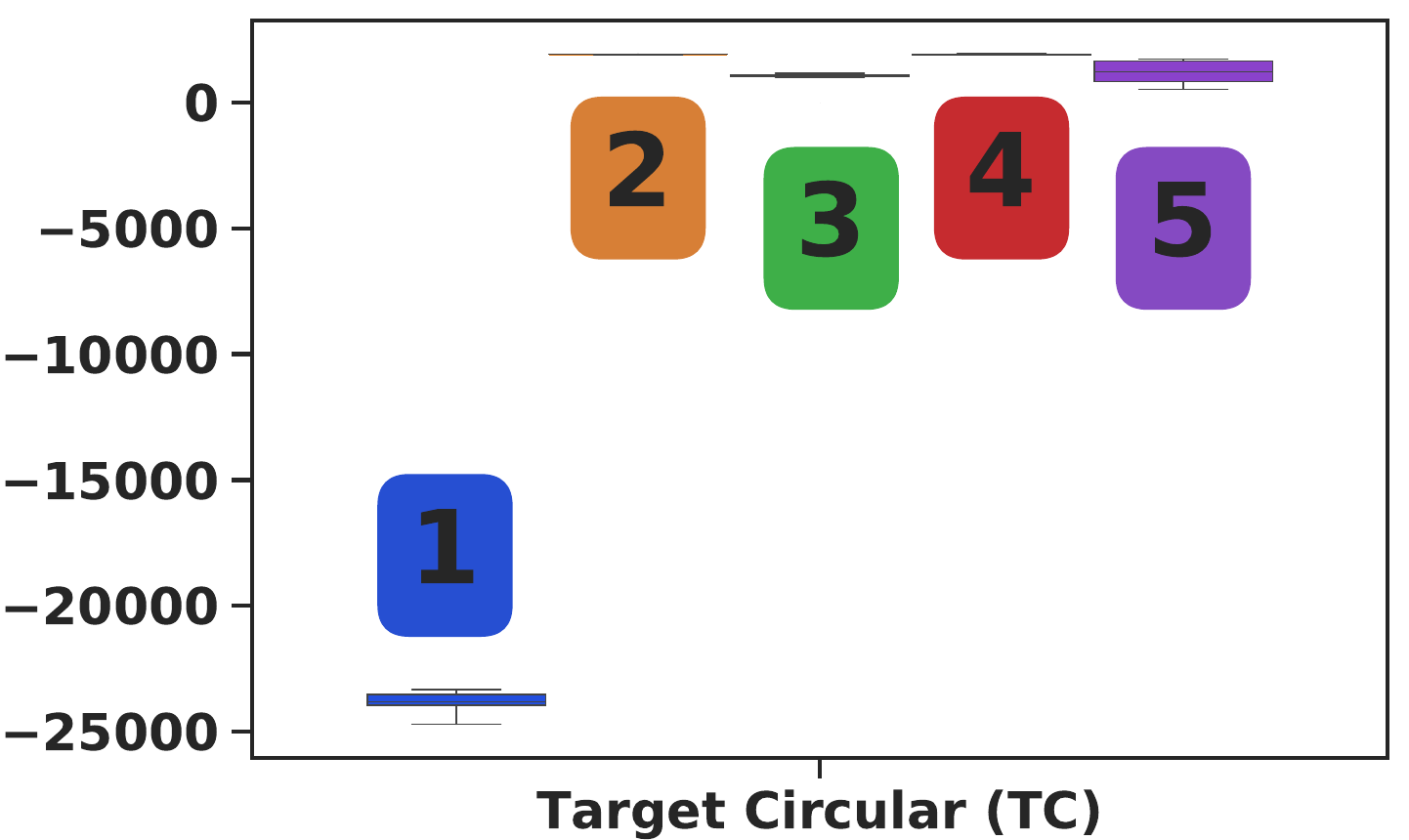}
    \includegraphics[width=0.31\linewidth]{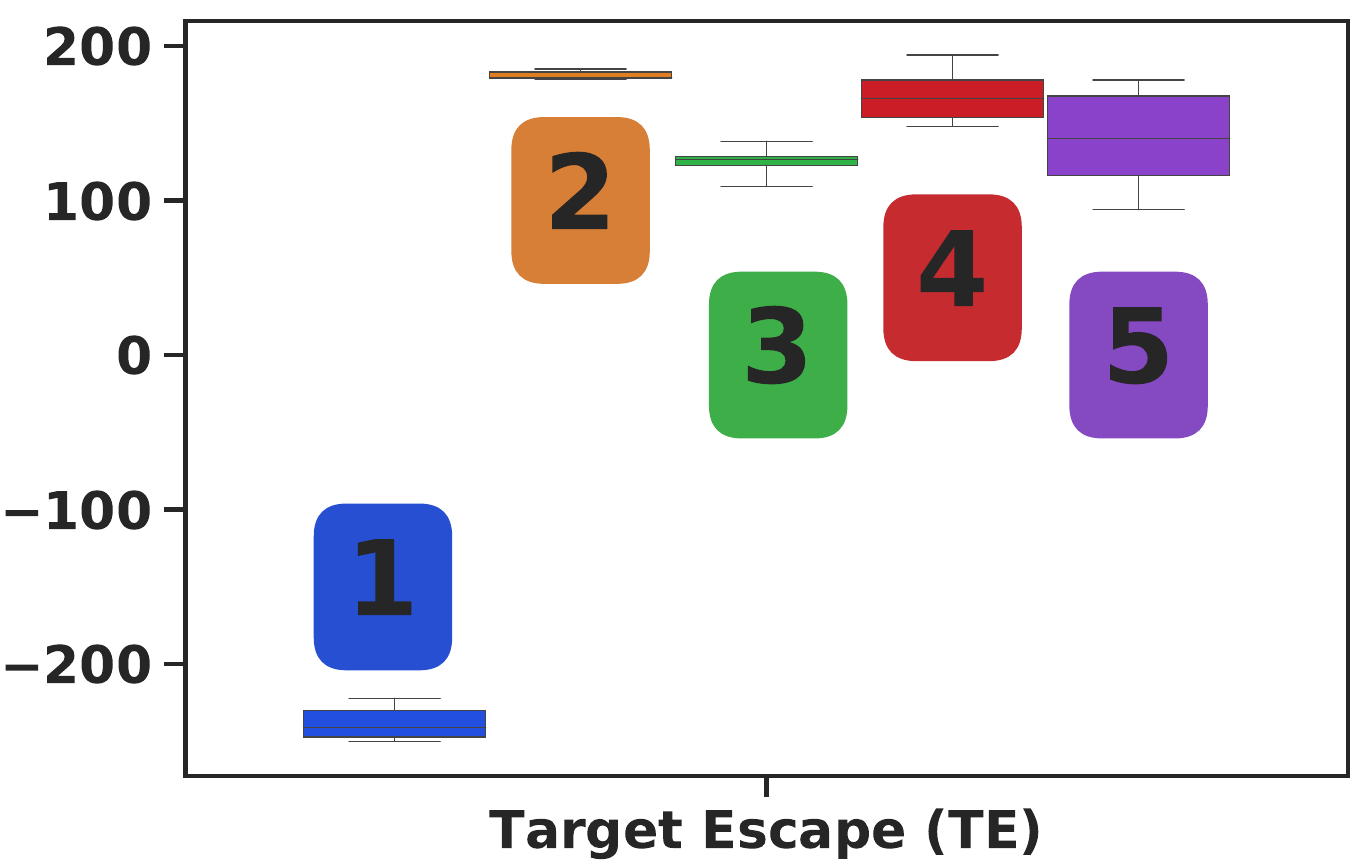}
    \includegraphics[width=.99\linewidth]{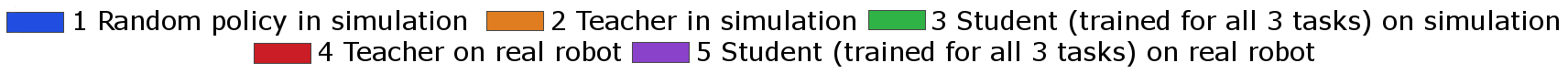}
    \caption{Main result: distillation in a continual learning setting of three teacher policies into a single student policy. The resulting policy is able to perform all three tasks both in simulation and in the real world, while minimizing forgetting.} 
    \label{fig:final_perf}
\end{figure}

We present our final results in Fig. \ref{fig:final_perf}. We used \textit{on policy} data generation and training using KL divergence loss with $\tau = 0.01$. 
We show box plots over 10 episodes of reward performances for teacher policies in each task, and for the distillation of the same three teachers into a single student using DisCoRL. Each policy is evaluated in simulation and also in real-life on the robot. As a reference, we also show the performance of a random agent in each task. Our approach is effective in a continual reinforcement learning setting: the performance of teachers and student are similar.

More precisely, there are two main challenges to overcome in our setting: learning a behaviour via distillation by using only a limited number of examples, and the reality gap which can notoriously \cite{tobin2017domain} introduce variations that may lead the policy to fail. Fig. \ref{fig:final_perf} demonstrates the efficiency of our approach at overcoming both of these issues: only a small fraction of performance is lost from teacher to student, and from simulation to reality. We can see that the single student distilled policy achieves close to maximum rewards in all tasks, in real-life.

\subsection{Negative results}
\label{subsec:neg_res}

While distillation is effective for policy transfer, we also tested other alternatives worth mentioning. 

\textbf{Elastic Weight Consolidation (EWC)} \cite{kirkpatrick2017overcoming} was implemented as a continual learning baseline to compare with the distillation method. EWC has the appealing advantage of not re-using any data from previous tasks. However, in all cases we found the method unsuccessful. 

Tuning the $\lambda$ parameter that controls the trade-off between weight protection and learning the new task showed that either $\lambda$ is too low and catastrophic forgetting happens, or $\lambda$ is too high and nothing new is learned (i.e., the full network is frozen).
A $\lambda$ value providing a proper balance in between both effects could not be found for such sequential tasks to be learned. 

\textbf{Progress and Compress (P\&C)} \cite{schwarz2018progress} was tested but as EWC, we had problems with the importance factor $\lambda$ and we where not able to learn three policies into a single model with this method.

% tim : it looks strange to me, maybe task label did not help for distillation but they could have been used in another
\textbf{Adding task labels for distillation.} Even if all tasks contain a visually differentiating identifier, they remain visually similar. In cases, we found that a distilled policy trained to perform well on several tasks can mix up tasks and thus not perform adequately. Hence, either adding tasks labels directly, or adding a module in the network that predicts the task label could be a way to improve the efficiency of distillation. However, none of the approaches were successful in practice, yielding the same results with or without task labels. 

\textbf{Gumbel-Softmax action sampling for the student} \cite{jang2016categorical}. This trick allows to sample from a categorical distribution using a softmax output layer. It has proven to be useful for action sampling in policy learning \cite{schulman2017proximal}. However, in our case we saw no improvement over a simple argmax strategy for action sampling when we used it on the student policy at test time.

%%%%%%%%%%%%%%%%%%%%%%%%%%%%%%%%%%%%%%%%%%%%%%%%%%%%%%%%%%%%%%%%%%%%%%%%%%%%%%%%
%%%%%%%%%%%%%%%%%%%%%%%%%%%%%%%%%%%%%%%%%%%%%%%%%%%%%%%%%%%%%%%%%%%%%%%%%%%%%%%%
%%
%                                    Discussion and Future Work
%%
%%%%%%%%%%%%%%%%%%%%%%%%%%%%%%%%%%%%%%%%%%%%%%%%%%%%%%%%%%%%%%%%%%%%%%%%%%%%%%%%
%%%%%%%%%%%%%%%%%%%%%%%%%%%%%%%%%%%%%%%%%%%%%%%%%%%%%%%%%%%%%%%%%%%%%%%%%%%%%%%%

\section{Discussion and Future Work}
\label{ref:discussion}

Continual learning is a complex field: every setting is different and expectations may vary from one algorithm to another. For example it is not easy to compare results with and without task indicator. Task labels always add information to learn or test, and thus, they often improve results. However in a realistic setting they may be lacking. 

Otherwise, the scalability vs stability trade-off is a difficult question. Learning online in a single model or a dual architecture scales well to a high number of tasks. However, this solution is often unstable, in particular because if a task fails, there is high risk of forgetting everything that has been learned previously. For example, in generative replay, the generator is used as a memory. However, if at some moment it diverges while learning, all data from the past is destroyed. 
The approach we propose uses soft-labelled samples as a memory, similarly to rehearsal methods, which will grow the memory continually. However, this brings no risk of forgetting or destroying past knowledge. 

Nevertheless, even if we believe this work proposes a stable and scalable framework for continual reinforcement learning, several possibilities for improvement exists. Our road map includes having not only a policy learned in a continual way, but also the SRL model associated. We would need to update the SRL model as new tasks are presented sequentially. One possible approach would be to use Continual SRL methods like S-TRIGGER \cite{caselles2019s} or VASE \cite{achille2018life}. Moreover, we would like to optimize more the memory needed to save samples by reducing their number and their size.

Our empirically best distillation strategy currently consists on performing early stopping on the joint policy $\pi_{d:1,..,n}$. In future work, we plan to perform automatic model selection for the final policy.

Finally, training policies on real robot experiences without the use of simulation would be desirable. However, at the moment, this is more a RL challenge than a CL challenge. One promising approach would be to use model-based RL while learning the SRL model to improve sample efficiency. Though, nowadays approaches still do not offer solutions working in a reasonable amount of time.

\section{Conclusion}

In this paper we presented DisCoRL, an approach for continual reinforcement learning.
The method consists of summarizing sequentially learned policies into a dataset to distill them into a student model.  It allows to learn sequential tasks in a stable pipeline without forgetting. Some loss in performance may occur while transferring knowledge from teacher to student, or while transferring a policy from simulation to real life.
Nevertheless, our experiments show promising results in simulated environments and real life settings.
Future work will evaluate it in more complex tasks.  

\section{Acknowledgement}
This work is supported by the EU H2020 DREAM project (Grant agreement No 640891). 

\clearpage

\bibliographystyle{apalike}
\bibliography{references.bib}  % .bib

\appendix
\section{Dataset generation}

The notation summarizing our data generation and distillation processes is:
\begin{itemize}
\item $Env_i$: Environment for task $i$.
\item $\mathcal{T}_i$: The set of all encountered environments when encountering task $i$.
\item $D_{R,i}$: Dataset generated by a random policy on environment $i$.
\item $E_i$: Encoder $i$ from SRL step from task $i$.
\item $\pi_i$: Policy $i$.
\item $D_{\pi_i}$: Dataset generated by and on-policy $\pi_i$.
\item $\pi_{d:1,...,i}$: Policy distilled on environment 1 to $i$.
\end{itemize}

While generating on-policy datasets $D_{\pi 1}$ (see Section 5.1.2) %\ref{subsec:oneTask})
for task 1 (TR), we allow the robot to perform a limited number of contacts with the target to reach ($N_{contacts}=10$) in order to mainly preserve the frames associated with the correct reaching behaviour. There are no such additional constraints when recording for task 2 (TC) or 3 (TE), the limit is the standard episode length, i.e. 250 time-steps.

\section{Evaluation of each task separately}
\label{ap:distill-sep}

Before moving to a continual setup, we should test if it is possible to solve each task separately using RL.

One of the challenges in CL is not the ability to distill knowledge, but the ability to know when a policy has been distilled, i.e. properly learned by the student. Due to the hypothesis of real continual learning settings where access to previous environments is not possible, one of the challenges is finding a proxy task that can help indicate when early stopping of the policy distillation can be applied. That proxy task or signal should be different from the reward achieved in previous environments, which is no longer available. In our experiments, we found empirically that a small number of epochs, i.e $N=4$, guarantee policy learning.

\newpage
\begin{figure}[h]
    \centering
    \includegraphics[scale=0.35]{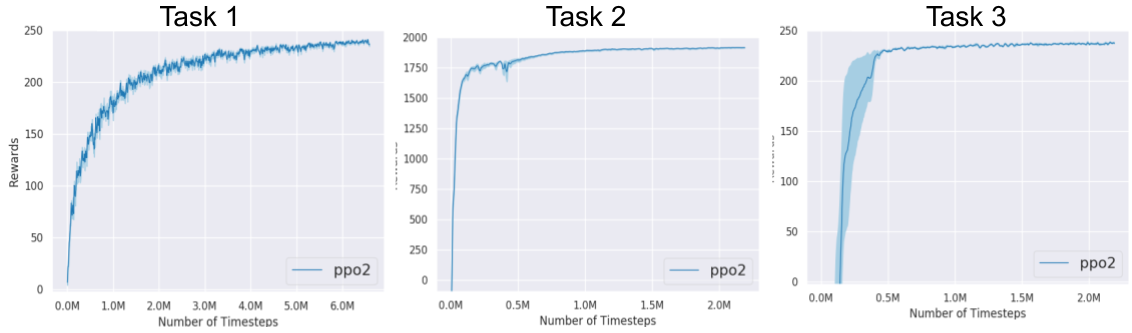}
    \caption{Mean and standard error of rewards during RL learning of each task separately. Each task is learned using the same type of SRL model (SRL Combination), trained on each environment. All three tasks are mastered within roughly 2M timesteps.}
    \label{fig:teacher_policies}
\end{figure}

\section{Memory needs}
The memory occupation by the different models that form part of our pipeline are the following:
\begin{itemize}
  \item \textbf{Dataset}: $D_{\pi_{1,2,3}}$: 554.6 MB
  \item \textbf{SRL model + RL policy} (PPO2): 4.8 MB + 0.143 MB 
  \item \textbf{RL model from raw pixels} : 0.143 MB
  \item \textbf{Distilled student policy} (replay model): 1.1 MB
\end{itemize}

\section{Evaluation of sequential policy learning via fine-tuning}
\label{ap:eval_forget}

\begin{figure}[hb]
    \centering
    \includegraphics[scale=0.15]{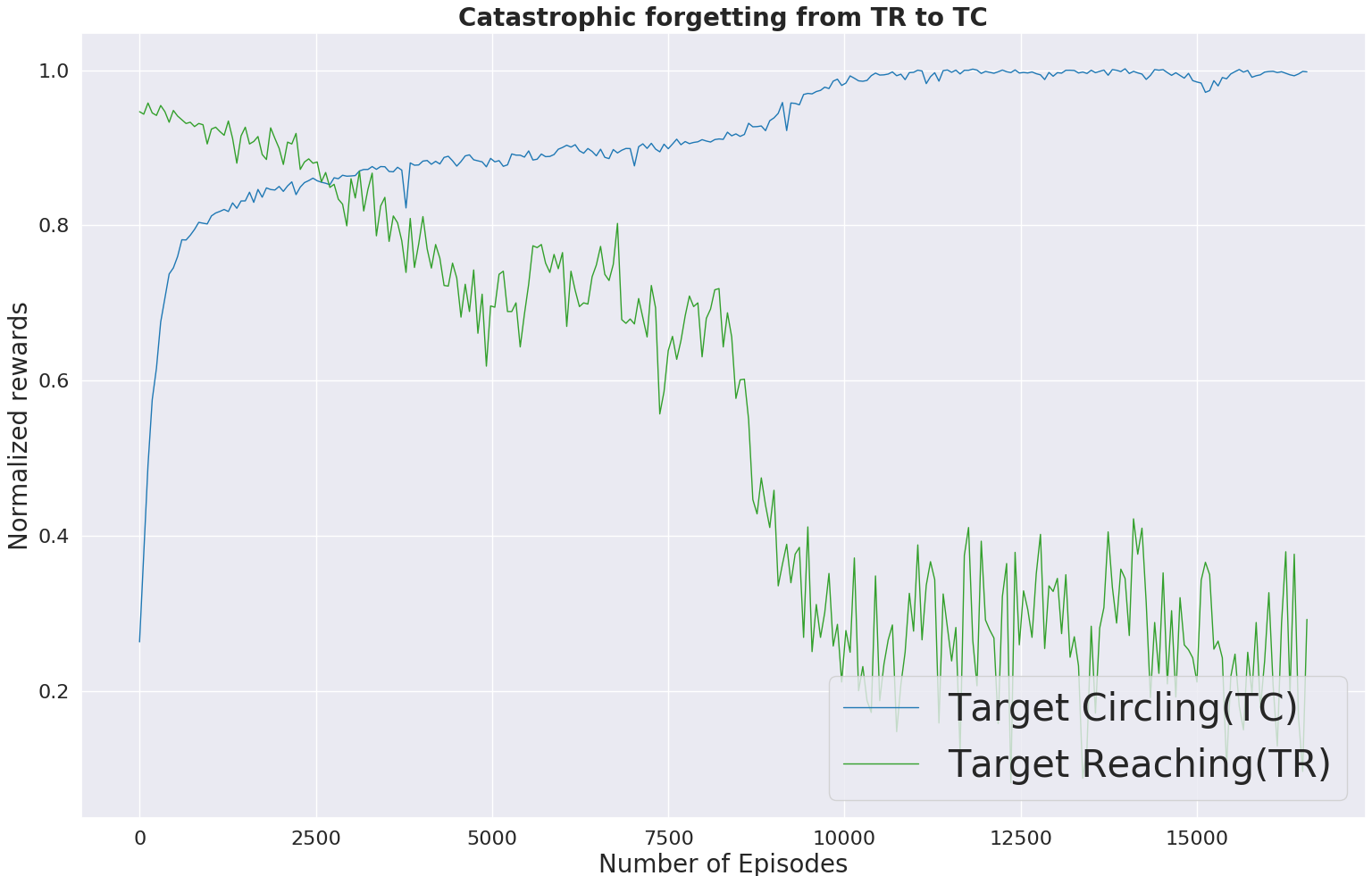}
    \caption{Demonstration of catastrophic forgetting phenomenon occurring while fine-tuning a network on a new task. The blue curve shows the progression of the mean reward of the policy currently being learned (task $i$: TC). The green curve represents its mean reward when evaluated on the task $i-1$ previously learned (first task: TR). Both evaluations were made with 5 random seeds.}
    \label{fig:cf-tr-2-rc}
\end{figure}

Results on policy fine-tuning (see Fig. \ref{fig:cf-tr-2-rc}) show that catastrophic forgetting occurs on a previously learned task $k < i$ (TR) when the fine-tuned policy reaches near convergence on a new task $i$ (TC). Moreover, it appears that the fine-tuning could be interrupted such that the resulting policy would have reasonable performances on the set of all encountered environments so far $\mathcal{T}_i=\{TC,TR\}$.

\section{Evaluating distillation while learning the knowledge to distill from}
\label{ap:distill-1-task}

\begin{figure}
    \centering
    \includegraphics[scale=0.25]{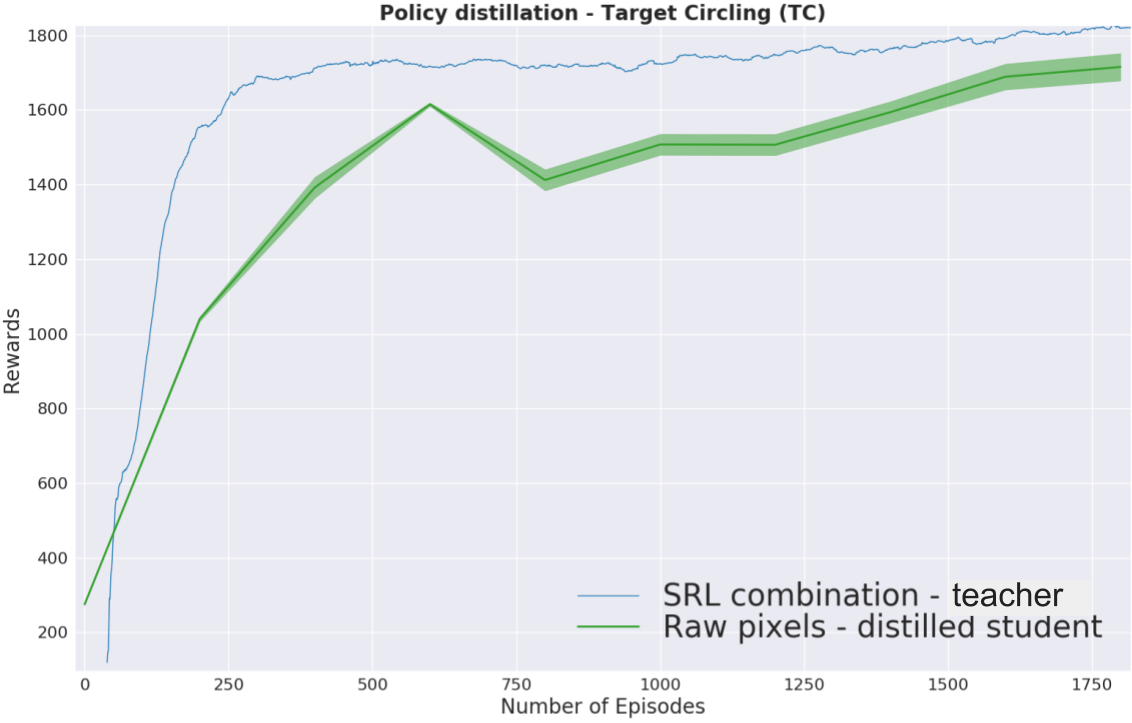}%distilled_from_single_CC.png}
    \caption{Demonstration of the effectiveness of distillation. Blue: RL training curve of an SRL based Policy (SRL Combination) on the target circling (TC) task. Green: Mean and standard deviation performance on 8 seeds of distilled student policy.  %\cnat{what is each point? timestep? episode? replace being concrete}
    The teacher policy in blue is distilled into a student policy every 200 episodes  (1 episode = 250 timesteps). 
    }
    \label{fig:distillation_cc}
\end{figure}

We performed a more explicit evaluation of distillation in the task 2 (Target Circling (TC)). While we train a policy using RL, we save the policy every 200 episodes (50K timesteps), and distill it into a new student policy which we test. This is illustrated in Fig. \ref{fig:distillation_cc}. Both curves are very close, which indicates that policy distillation enables to reproduce the skills of a teacher policy regardless of the teacher's state of convergence on the evaluated task. Moreover, distillation is able to transfer knowledge from teacher policy into a student using a small number of observations, i.e only 15k samples (w.r.t. the volume of samples required to learn the teacher policy, see Fig. \ref{fig:teacher_policies}).

\section{State Representation Learning (SRL) model}

\begin{figure}[ht!]
    \centering
    \includegraphics[scale=0.25]{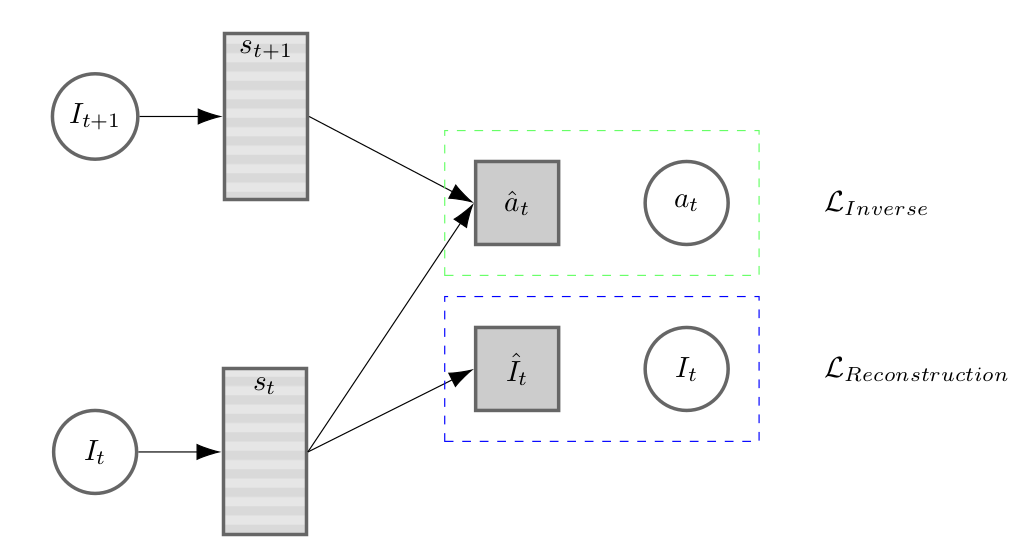}

\caption{\textit{SRL Combination} model: combines the prediction of an image $I$'s reconstruction loss and an inverse dynamics model loss in a state representation $s$. Arrows represent inference, dashed frames represent losses computations, rectangles are state representations, circles are real observed data, and squares are model predictions; $t$ represents the timestep.}

\label{fig:split-model}
\end{figure}

\newpage
\section{Distillation model architecture}

Architecture available at : \url{https://github.com/araffin/srl-zoo/blob/438a05ab625a2c5ada573b47f73469d92de82132/models/models.py#L179-L214}

\end{document}